\documentclass[11pt,twoside,twocolumn,a4paper]{article}

\usepackage{acvrw}
\usepackage{times}
\usepackage{epsfig}
\usepackage{graphicx}
\usepackage{amsmath}
\usepackage{amssymb}
\usepackage{hyperref}
\usepackage{subcaption}
\usepackage{sidecap}

\newcommand{\furl}[1] {\footnote{\url{#1}}}
\graphicspath{{images/}}

\acvrwfinalcopy 

\def\acvrwPaperID{15} 

\ifacvrwfinal\fi
\begin{document}

\title{The Problem of Fragmented Occlusion in Object Detection}

\author{Julian Pegoraro\textsuperscript{1}, Roman Pflugfelder\textsuperscript{1,2} \\
\textsuperscript{1}~AIT Austrian Institute of Technology, \textsuperscript{1,2}~TU Wien\\
{\tt\small {\{julian.pegoraro|roman.pflugfelder\}}@ait.ac.at}, \tt\small {roman.pflugfelder@tuwien.ac.at}
}

\maketitle
\ifacvrwfinal\thispagestyle{fancy}\fi


\begin{abstract}
Object detection in natural environments is still a very challenging task, even though deep learning has brought a tremendous improvement in performance over the last years. A fundamental problem of object detection based on deep learning is that neither the training data nor the suggested models are intended for the challenge of fragmented occlusion. Fragmented occlusion is much more challenging than ordinary partial occlusion and occurs frequently in natural environments such as forests. A motivating example of fragmented occlusion is object detection through foliage which is an essential requirement in green border surveillance. This paper presents an analysis of state-of-the-art detectors with imagery of green borders and proposes to train Mask R-CNN on new training data which captures explicitly the problem of fragmented occlusion. The results show clear improvements of Mask R-CNN with this new training strategy (also against other detectors) for data showing slight fragmented occlusion.
\end{abstract}


\section{Introduction}
Automated surveillance at green borders has become a hot topic for European border guards. Border guards today face several challenges in protecting EU borders. One well known occasion in public is illegal migration which had its peak in 2015.

Border surveillance today limited to 2D imaging sensors consists of color and thermal cameras, mounted on poles or used as handheld cameras by the border guards. Innovating these technical systems by adding further capabilities of automatic inference such as the automatic detection of persons, vehicles, animals and suspicious objects in general will need to apply object detectors to such imagery. 

However, video of green borders especially at EU borders show significant differences to typical imagery of video surveillance such as indoor video or video taken in man-made outdoor scenes. For example, green borders are scenes showing dense forest, hills, harsh weather and climate conditions. Such scenes draw challenges to automated surveillance and raise several interesting research questions.

\begin{figure}
\centering
\includegraphics[width=0.42\columnwidth, height=4.5cm]{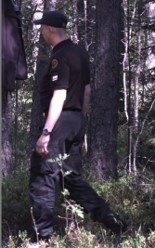}\vspace{1pt} \includegraphics[width=0.42\columnwidth, height=4.5cm]{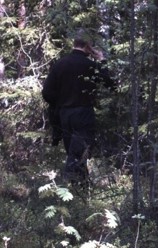}\\[1pt]
\includegraphics[width=0.42\columnwidth, height=4.5cm]{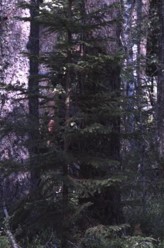}\vspace{1pt} \includegraphics[width=0.42\columnwidth, height=4.5cm]{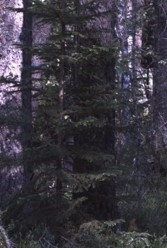}
\caption{The problem of fragmented occlusion in object detection. Top Left: no occlusion (level $L_0$). Top Right: slight occ. ($L_1$). Bottom Left: moderate occ. ($L_2$). Bottom Right: heavy occ. ($L_3$) occlusion.\label{fig:frag-occlusion}}
\end{figure}

This paper considers a challenge for state-of-the-art object detection in green border surveillance which is the problem of through foliage detection. To the best of our knowledge, none of the current approaches for object detection allow the detection of objects through foliage. This problem raises an interesting scientific question, namely how to detect objects with fragmented occlusion? This problem is also different to the problem of partial occlusion in object detection. Fragmented occlusion occurs by viewing objects behind tree ans bush leaves. Contrary to partial occlusion, fragmented occlusion gives no clear view on minimal recognisable parts of the object~\cite{ullman-pnas2016} which is used to detect the object~\cite{nebehay-cvpr2015}.

We show in this work that the state-of-the-art in object detection fails on fragmented occlusion even for the moderate case. For this, we created a new dataset~(Figure~\ref{fig:frag-occlusion}) capturing people behind trees. We labelled nearly 40,000 images in three representative videos. This data raises new challenges on the labelling and evaluation which we only partially answer in this paper. For example, bounding boxes are the standard in current evaluation of detectors but such labels are hard to find in data that contains fragmented occlusion. As the state-of-the-art detectors deliver bounding boxes, fragmented occlusion poses new questions on the evaluation methodology.

Furthermore, we augmented Microsoft COCO\furl{http://cocodataset.org} training data by occluding the ground truth masks similarly as leaves occlude people behind bushes and trees. We then show results on training Mask R-CNN~\cite{MaskR-CNN} on this new data showing improvement of Mask R-CNN trained on the original data with slight fragmented occlusion.

\section{Related Work}
State-of-the-art object detection is based on deep learning. Two-stage detectors work by finding as an intermediate step bounding box proposals~\cite{R-CNN, FastR-CNN} on the feature maps of the backbone CNN. A region proposal network further improves efficiency~\cite{FasterR-CNN, MaskR-CNN}. One-stage detectors regress the bounding boxes directly~\cite{YOLOv3, SSD} which is computationally efficient on GPUs but this approach is inherently less accurate as it assumes a coarsely discretised search space. Although these methods show usually excellent performance for fully visible objects, they break down in the case of fragmented occlusion. Fragmented occlusion has not been considered for object detection so far, however there is literature about this topic in the field of motion analysis~\cite{black-CVIU1996}.

\section{Methodology}
We created a dataset recorded in a forest consisting of three videos with a total of 18,360 frames and 33,933 bounding boxes which were manually defined by human annotators. These bounding boxes are divided into four different occlusion levels including the unoccluded case (Figure~\ref{fig:frag-occlusion}).

\begin{figure}
\centering
\includegraphics[width=0.42\columnwidth]{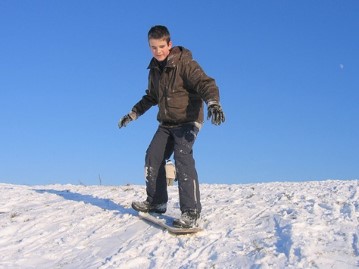}\vspace{1pt} \includegraphics[width=0.42\columnwidth]{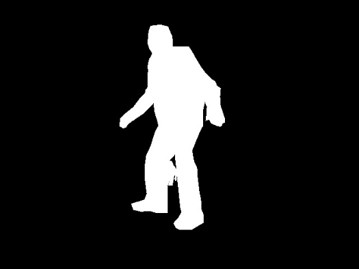}\\[1pt]
\includegraphics[width=0.42\columnwidth]{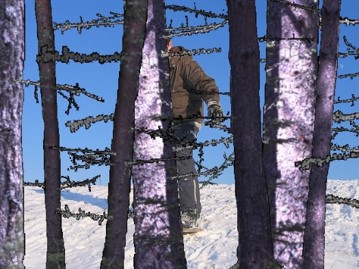}\vspace{1pt} \includegraphics[width=0.42\columnwidth]{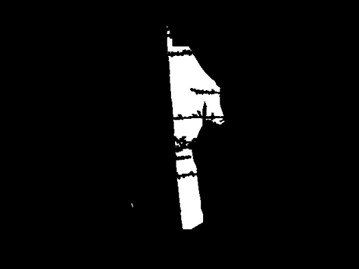}
\caption{A training image from Microsoft COCO (\url{http://images.cocodataset.org/train2017/000000001700.jpg}). Top Left: the image. Top Right: Segmentation mask of the image. Bottom Left:  image overlaid with artificial trees. Bottom Right: Mask of the overlaid image. \label{fig:CocoImage}}
\end{figure}

Then, we extended the Microsoft COCO dataset by adding artificial trees as foreground to the images of objects (Figure~\ref{fig:CocoImage}). We chose this dataset, because it contains pixel-wise segmentation masks in the ground truth as well as a large number of different categories including the human person.

The underlying basic idea of our approach is to add artificial fragmented occlusion to Microsoft COCO and train Mask R-CNN on this new data. By this we can adapt the original distribution of data to the case of fragmentally occluded objects. Since we are only interested in humans, we apply this augmentation only to images containing humans and use only these images for training. The trees used for the augmentation are generated from real images we have obtained from the test data. The method generates whole artificial trees by randomly adding branches to previously manually segmented tree trunks. In total 14 such trunks are extracted from the test dataset. The branches attached to these trunks are also randomly generated by also adding a few manually segmented leaves.

The trees are placed in front of objects by randomly selecting the x-coordinate on which they will be placed and an angle at which the tree will be rotated. The calculated foreground is applied to the image and its negative mask is multiplied by the segmentation mask of the objects in the image. The Mask R-CNN model is then trained with the augmented images. The selected backbone model is the Inception v2 \cite{InceptionV2} network. This network is selected for its faster computation.




\section{Evaluation}
\label{sec:evaluation}
To evaluate whether training with the augmented dataset is useful, the model trained on the augmented data must be compared with the model not trained on this data. However, the intersection over Union (IoU) measure is not meaningful in this case.

Standard evaluation metrics such as the mean average precision (mAP) define an IoU threshold (e.g. 0.5) and check whether a ground truth object and a detected object have an IoU value above this value. If this is the case, the detected object is defined as a True Positive (TP). If an object is detected but there is no respective ground truth with an IoU above this specific threshold, the detected object is defined as a False Positive (FP). If there is ground truth but no detected object with an IoU above the threshold, the object is defined as a False Negative (FN).

\begin{figure}
	\centering
	\includegraphics[width=0.85\columnwidth]{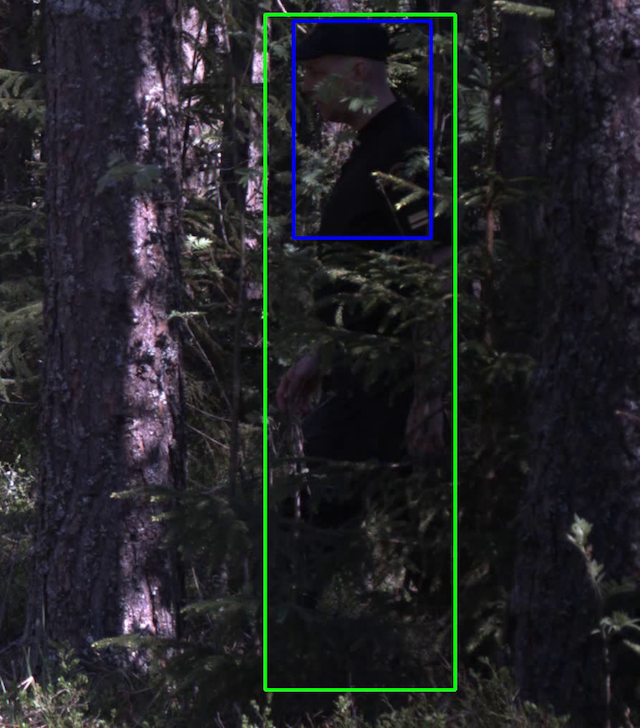}
	\caption{Ground truth (green) and the detection (blue) vary substantially due to the occlusion effects.}
	\label{fig:headdet}
\end{figure}

These evaluation methods cannot be easily applied to ground truth showing fragmented occlusion, because of the following two observations:

\textbf{IoU too small:} Since the data is based on fragmented detections, a detector can only detect parts of the person. An image where this problem occurs is shown in Figure~\ref{fig:headdet}. The bounding box is clearly a TP, based on the fact, that fragmented objects should be detected, but due to the occlusion by the branches of the tree, the whole body cannot be recognized. This leads to an IoU of only $\approx 0.2$.

\begin{SCfigure*}
\centering
\includegraphics[height=5cm]{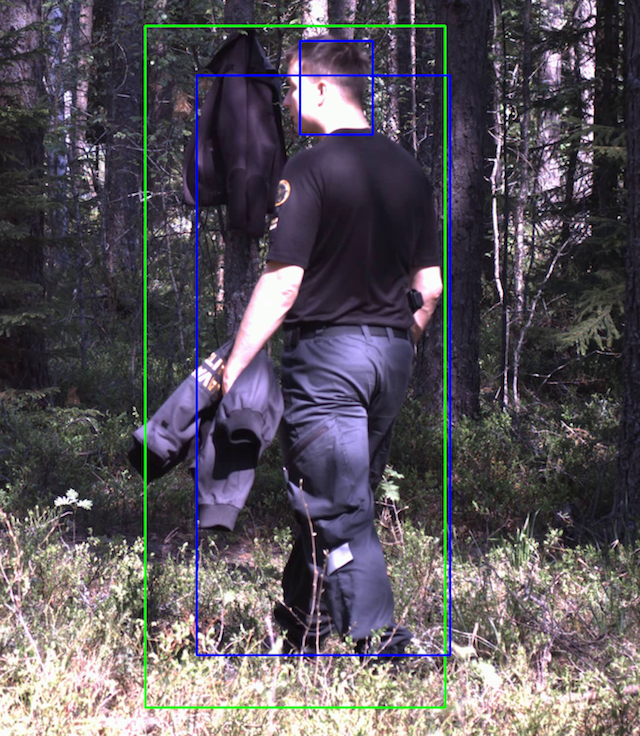} \quad \includegraphics[height=5cm]{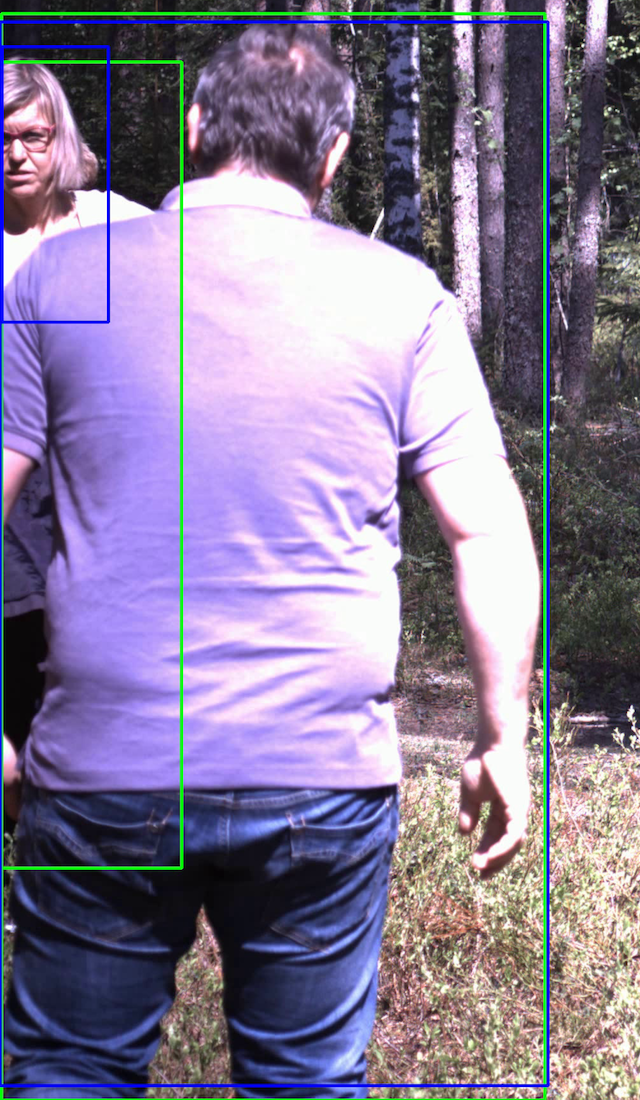} \quad  \includegraphics[height=5cm]{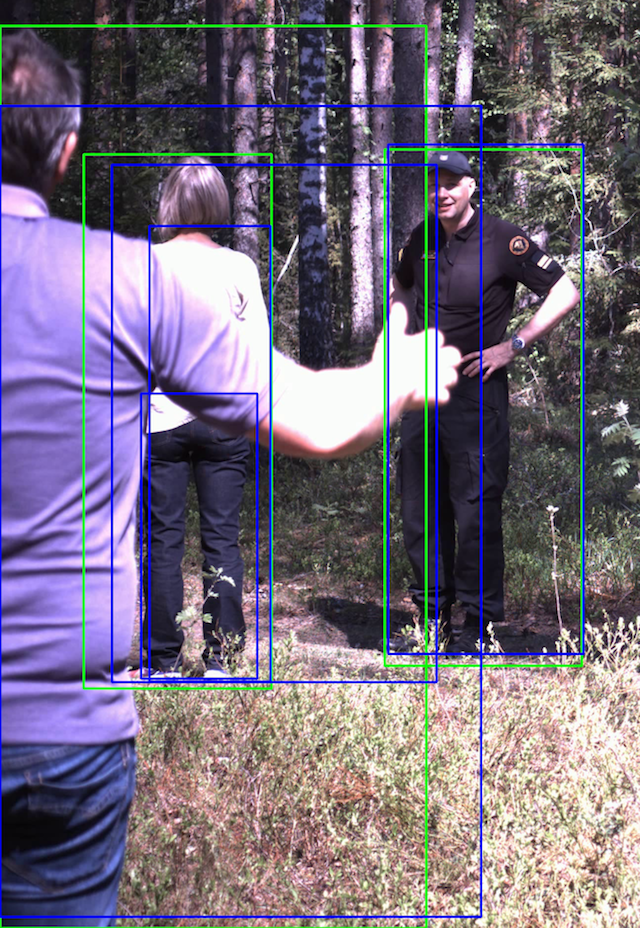}
\caption{The problem of multiple detections. Ground truth is shown in green. Left: state-of-the-art yields two bounding boxes of the same, single person. Middle: two persons are visible. Detection yields two bounding boxes which are diffucult to associate. Right: an even harder case with three persons.}
\label{fig:multiple}
\end{SCfigure*}

\textbf{Multiple detections:} Another major problem with the standard evaluation metrics is that exactly one detected bounding box and one ground truth bounding box match. However, when handling fragmented objects, human heads and/or other body parts should be detected separately if body parts are covered. This creates the problem that parts of the body (like a head) is detected as well as the whole body. Figure~\ref{fig:multiple} shows some examples.

To tackle these two problems, this paper proposes a different evaluation metric. For each bounding box in the evaluation data set, we calculate the maximum region in the image where there is no overlap with another ground truth bounding box. This region is then extracted and fed into the model. If the model detects an object, we define it as TP, otherwise as FN. To assess FPs, we create an additional dataset that represents the maximum region in an image without overlap with any ground truth bounding box. We extracted in total 45,340 such regions with different aspect ratios, different parts of the image and at different time instants. In addition to FPs, we can also calculate the TNs using this evaluation metrics.

\begin{figure}
	\centering
	\includegraphics[width=\columnwidth]{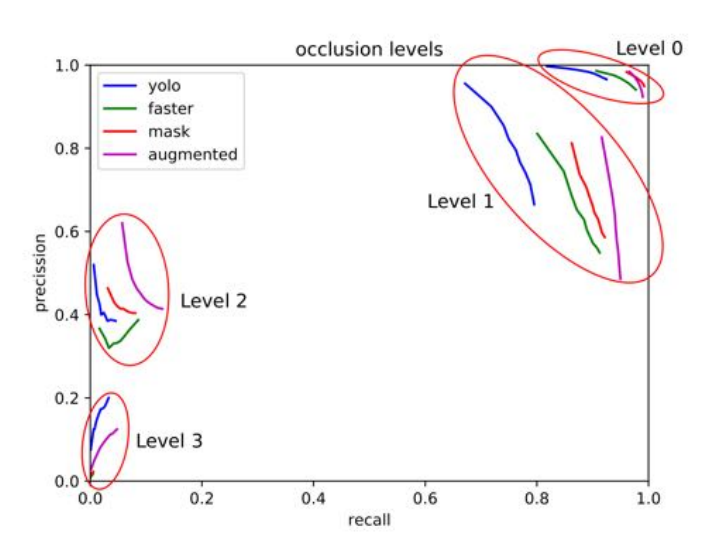}
	\caption{This ROC plot shows results of Faster R-CNN (green), YOLO (blue), Mask R-CNN (red) and our method (purple) for all occlusion levels.}
	\label{fig:results}
\end{figure}

Figure~\ref{fig:results} shows these results as recall vs. precision curve (ROC). There is no significant difference between Mask R-CNN trained on Microsoft COCO and on the augmented dataset for $L_0$ occlusion. However, clear improvement has been achieved for $L_1$ and $L_2$ occlusion which proves the applicability of the idea to model fragmented occlusion by the masks. Nevertheless, all approaches basically do not reach the expected robustness and accuracy for moderate $L_2$ and heavy $L_3$ occlusion. One reason for this is that our current technique is not accurate enough to model fragmented occlusion. Furthermore, clear limits exist as heavy fragmented occlusion removes local spatial and structural information necessary for current approaches in object detection.

We further recognise that bounding box labelling is not the appropriate approach for labelling data showing fragmented occlusion. Especially for $L_3$ and $L_4$ occlusion, it is frequently impossible to manually define the bounding box. Such occlusion levels allow an approximate localisation of the object in the image but make the observation of the object's extent impossible. While the recall in Figure~\ref{fig:results} is still meaningful, the precision is basically undefined. This observation has severe consequences on the labelling, but also on the evaluation and on the detector which we leave open for future research.


\section{Conclusion}
This paper formulates a new scientific question on object detection with fragmented occlusion which is different to partial occlusion. We show by a study that current object detectors fail in this case. We generated and labelled a new dataset showing people behind trees in a forestry environment. Such scenes frequently occur in border surveillance which has become very important in EU security policies. We try to tackle the occlusion challenge by augmenting Microsoft COCO including the pixel-wise segmentation masks to capture the occlusion problem. We show that Mask R-CNN trained on this data improves on fragmented occlusion, however, we also observe severe loss of spatial, structural information and that the bounding box itself is not the appropriate description to cope with fragmented occlusion. This has severe implications on the detection approach itself, but also on dataset labelling and evaluation. A potential solution is left open for future work.
\section*{Acknowledgments}
This research was supported by the European Union H2020 programme under grant agreement FOLDOUT-787021. We thank all our students on internship to label the new dataset.

{\small
\bibliographystyle{ieee}
\bibliography{acvrw}
}

\end{document}